\documentclass[preprint,5p]{elsarticle}

\usepackage{prletters}

\usepackage{amsmath}
\usepackage{amsthm}
\usepackage{amsbsy}
\usepackage{amssymb}
\usepackage{dsfont}
\usepackage{nicefrac}
\usepackage{siunitx}
\usepackage{etoolbox}
\usepackage{xcolor}
\usepackage{epigraph}
\usepackage{arydshln}
\usepackage[first=0, last=9]{lcg}
\usepackage{booktabs}
\usepackage[normalem]{ulem}

\newcommand{\argmin}{\operatornamewithlimits{argmin}}
\newcommand{\argsoftmax}{\operatornamewithlimits{softmax}}

\renewcommand{\bfseries}{\fontseries{b}\selectfont}
\newrobustcmd{\B}{\bfseries}

\newcommand*{\mynum}[1]{\num[output-decimal-marker={.},
                             round-mode=places,
                             round-precision=1,
                             group-digits=false]{#1}}

\newcommand{\doi}[1]{~\mbox{doi}:\href{https://doi.org/#1}{\texttt{#1}}}
\newcommand{\shboldres}[2]{\textbf{\shnormres{#1}{#2}}}
\newcommand{\shnormres}[2]{{\mynum{#1}}{\scriptsize$\pm$\mynum{#2}}}

\newcommand{\minispace}{

\vspace{0.07em}}

\usepackage{scalerel}
\usepackage{tikz}
\usetikzlibrary{svg.path}

\definecolor{orcidlogocol}{HTML}{A6CE39}
\tikzset{
    orcidlogo/.pic={
        \fill[orcidlogocol] svg{M256,128c0,70.7-57.3,128-128,128C57.3,256,0,198.7,0,128C0,57.3,57.3,0,128,0C198.7,0,256,57.3,256,128z};
        \fill[white] svg{M86.3,186.2H70.9V79.1h15.4v48.4V186.2z}
        svg{M108.9,79.1h41.6c39.6,0,57,28.3,57,53.6c0,27.5-21.5,53.6-56.8,53.6h-41.8V79.1z M124.3,172.4h24.5c34.9,0,42.9-26.5,42.9-39.7c0-21.5-13.7-39.7-43.7-39.7h-23.7V172.4z}
        svg{M88.7,56.8c0,5.5-4.5,10.1-10.1,10.1c-5.6,0-10.1-4.6-10.1-10.1c0-5.6,4.5-10.1,10.1-10.1C84.2,46.7,88.7,51.3,88.7,56.8z};
    }
}

\newcommand\orcidicon[1]{\href{https://orcid.org/#1}{\mbox{\scalerel*{
\begin{tikzpicture}[yscale=-1,transform shape]
\pic{orcidlogo};
\end{tikzpicture}
}{|}}}}

\usepackage{hyperref}
\usepackage{enumitem}

\begin{document}






 
 






\begin{frontmatter}

\title{Continual Semi-Supervised Learning through Contrastive Interpolation Consistency}

\author[a1]{Matteo~Boschini}
\ead{matteo.boschini@unimore.it}

\author[a1]{Pietro~Buzzega}
\ead{pietro.buzzega@unimore.it}

\author[a1]{Lorenzo~Bonicelli\corref{cor1}}
\cortext[cor1]{Corresponding author}
\ead{lorenzo.bonicelli@unimore.it}

\author[a1]{Angelo~Porrello}
\ead{angelo.porrello@unimore.it}

\author[a1]{Simone~Calderara}
\ead{simone.calderara@unimore.it}

\address[a1]{University of Modena and Reggio Emilia, Via Vivarelli 10, Modena, Italy}

\begin{abstract}
Continual Learning (CL) investigates how to train Deep Networks on a stream of tasks without incurring \textit{forgetting}. CL settings proposed in literature assume that every incoming example is paired with ground-truth annotations. However, this clashes with many real-world applications: gathering labeled data, which is in itself tedious and expensive, becomes infeasible when data flow as a stream.
This work explores \textit{Continual Semi-Supervised Learning} (CSSL): here, only a small fraction of labeled input examples are shown to the learner. We assess how current CL methods (\textit{e.g.}: EWC, LwF, iCaRL, ER, GDumb, DER) perform in this novel and challenging scenario, where overfitting entangles forgetting. 
Subsequently, we design a novel CSSL method that exploits \textit{metric learning} and \textit{consistency regularization} to leverage unlabeled examples while learning. We show that our proposal exhibits higher resilience to diminishing supervision and, even more surprisingly, relying only on $25\%$ supervision suffices to outperform SOTA methods trained under full supervision.
\end{abstract}


\end{frontmatter}


\section{Introduction}
\label{sec:intro}
Perceptual information flows as a continuous stream, in which a certain data distribution may occur once and not recur for a long time. Unfortunately, this violates the i.i.d.\ assumption at the foundation of most Deep Learning algorithms and leads to the catastrophic forgetting~\cite{mccloskey1989catastrophic} problem, where the acquired knowledge is rapidly overwritten by the new one. In practical scenarios, we would prefer a system that learns incrementally from the raw and non-i.i.d.\ stream of data, possibly ready to provide answers at any moment. The design of such lifelong-learning algorithms is the aim of Continual Learning (CL)~\cite{de2019continual}.

Works in this field typically test the proposed methods on a series of image-classification tasks presented sequentially. The latter are built on top of image classification datasets (\textit{e.g.}:~MNIST, CIFAR, etc.) by allowing the learner to see just a subset of classes at once. While these experimental protocols validly highlight the effects of forgetting, they assume that all incoming data are labeled. 

In some scenarios, this condition does not represent an issue and can be easily met. This may be the case when ground-truth annotations can be directly and automatically collected (\textit{e.g.}: a robot that explores the environment and learns to avoid collisions by receiving direct feedback from it~\cite{aljundi2019task}). However, when the labeling stage involves human intervention (as holds in a number of computer vision tasks such as classification, object detection~\cite{zhou2020lifelong}, etc.), relying only on full supervision clashes with the pursuit of lifelong learning. Indeed, the adaptability of the learner to incoming tasks would be bottlenecked by the speed of the human annotator: updating the model continually would lose its appeal w.r.t.\ the trivial solution of re-training from scratch. Therefore, we advocate taking into account the rate at which annotations are available to the learner.

To address this point, the adjustment of the prediction model can be simply limited to the fraction of examples that can be labeled in real-time. Our experiments show that this results in an expected degradation in terms of performance. Fortunately, the efforts recently made in \textit{semi-supervised learning}~\cite{olivier2006semi,tarvainen2017mean} come to the rescue: by revising these techniques to an incremental scenario, we can still benefit from the remaining part of the data represented by unlabeled observations.
We argue that this is true to the lifelong nature of the application and also allows for exploiting the abundant source of information given by unlabeled data.
\begin{figure*}[!t]
    \centering
    \includegraphics[width=.99\textwidth]{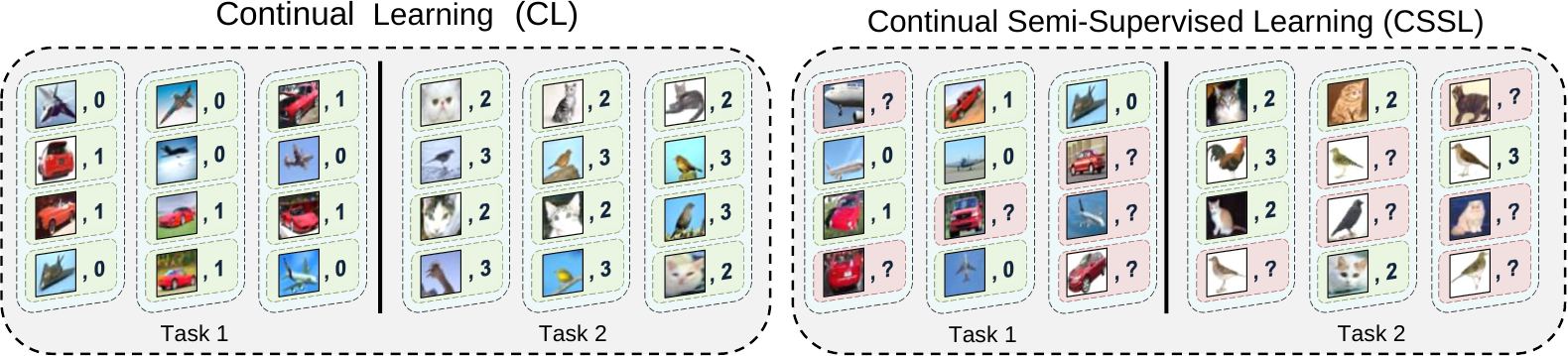}
    \caption{Overview of the Continual Semi-Supervised Learning (CSSL) setting. Input batches include both labeled (\textit{green}) and unlabeled (\textit{red}) examples.}
    \vspace{-0.7em}
    \label{fig:clls}
\end{figure*}
To sum up, our work incorporates the features described above in a new setting called \textbf{Continual Semi-Supervised Learning (CSSL)}: a scenario where just \textbf{one out of} $\boldsymbol{k}$ examples is presented with its ground-truth label. At training time, this corresponds to providing a ground-truth label for any given example with uniform probability~$\nicefrac{1}{k}$ (as shown in Fig.~\ref{fig:clls} for $k=2$). 

Taking one more step, our proposal aims at filling the gap induced by partial annotations: \textbf{Contrastive Continual Interpolation Consistency} (CCIC), which imposes consistency among augmented and interpolated examples while exploiting secondhand information peculiar to the Class-Incremental setting. Doing so, we grant performance that matches and even surpasses that of the fully-supervised setting. We finally summarize our contributions:
\begin{itemize}[noitemsep]
    \item We propose CSSL: a scenario in which the learner must learn continually by exploiting both supervised and unsupervised data at the same time;
    \item We empirically review the performance of SOTA CL models at varying label-per-example rates, highlighting the subtle differences between CL and CSSL;
    \item Exploiting semi-supervised techniques, we introduce a novel CSSL method that successfully addresses the new setting and learns with limited labels;
    \item Surprisingly, our evaluations show that full supervision does not necessarily upper-bound partial supervision in CL: $25\%$ labels can be enough to outperform SOTA methods using all ground truth.
\end{itemize}
\vspace{-1em}
\section{Related Work} \label{sec:related}
\subsection{Continual Learning Protocols}
Continual Learning is an umbrella term encompassing several slightly yet meaningfully different experimental settings~\cite{farquhar2018towards,van2019three}. \textit{Van de Ven et al.\ }produced a taxonomy~\cite{van2019three} describing the following three well-known scenarios. \textbf{Task-Incremental Learning (Task-IL)} organizes the dataset in tasks comprising of disjoint sets of classes. The model must only learn (and remember) how to correctly classify examples within their original tasks. \textbf{Domain-Incremental Learning (Domain-IL)} presents all classes since the first task: distinct tasks are obtained by processing the examples with distinct transformations (\textit{e.g.}:\ pixel permutations or image rotations) which change the input distribution. \textbf{Class-Incremental Learning (Class-IL)} operates on the same assumptions as Task-IL, but requires the learner to classify an example from any of the previously seen classes with no hints about its original task. Unlike Task-IL, this means that the model must learn the joint distribution from partial observations, making this the hardest scenario~\cite{van2019three}. For such a reason, we focus on limited labels within the Class-IL formulation. 
\minispace
\noindent\textbf{Towards realistic setups.} Several recent works point out that these classic settings lack realism~\cite{aljundi2019gradient} and consequently define new scenarios by imposing restrictions on what models are allowed to do while learning.
\textbf{Online Continual Learning} forbids multiple epochs on the training data on the grounds that real-world CL systems would never see the same input twice~\cite{lopez2017gradient,riemer2018learning,chaudhry2020using}.
\textbf{Task-Free Learning} does not provide task identities either at inference or at training time~\cite{aljundi2019gradient}. This is in contrast with the classic settings that signal task boundaries to the learner while training, thus allowing it to prepare for the beginning of a new task.

This work also aims at providing a more realistic setup: instead of focusing on model limitations, we acknowledge that requiring fully labeled data can hinder the extension of CL algorithms to real-time and in-the-wild scenarios.
\minispace
\noindent\textbf{Continual Learning with Unsupervised Data}. Some attempts have been recently made at improving CL methods by exploiting unlabeled data. \textit{Zhang et al.}\ proposed the \textbf{Deep Model Consolidation} framework~\cite{zhang2020class}; in it, a new model is first specialized on each new encountered task, then a unified learner is produced by distilling knowledge from both the new specialist and the previous incremental model. Alternatively, \textit{Lechat et al.}\ introduced \textbf{Semi-Supervised Incremental Learning}~\cite{lechat2021semi}, which alternates unsupervised feature learning on both input and auxiliary data with supervised classification. 

We remark that both these settings are significantly different from our proposed \textbf{CSSL} as we do not separate the supervised and unsupervised training phases. On the contrary, we intertwine both kinds of data in all drawn batches in varying proportions and require that the model learns from both at the same time. Additionally, we do not exploit auxiliary unsupervised external data to supplement the training set; instead, we reduce the original supervised data to a fraction, thus modeling supervision becoming available on the input stream at a much slower rate.
\subsection{Continual Learning Methods}
Continual Learning methods have been chiefly categorized in three families~\cite{farquhar2018towards,de2019continual}.
\minispace
\textbf{Architectural methods} employ tailored architectures in which the number of parameters dynamically increases~\cite{serra2018overcoming, fernando2017pathnet} or a part of them is devoted to a distinct task~\cite{mallya2018packnet}. While being usually very effective, they depend on the availability of task labels at prediction time to prepare the model for inference, which limits them to Task-IL.
\minispace
\noindent\textbf{Regularization methods} condition the evolution of the model to prevent it from forgetting previous tasks. This is attained either by identifying important weights for each task and preventing them from changing in later ones~\cite{kirkpatrick2017overcoming,zenke2017continual} or by distilling the knowledge from previous model snapshots to preserve the past responses~\cite{li2017learning, schwarz2018progress}. 
\minispace
\noindent\textbf{Rehearsal methods} maintain a fixed-size working memory of previously encountered exemplars and recall them to prevent forgetting~\cite{ratcliff1990connectionist}. This simple solution has been expanded upon in many ways, \textit{e.g.}\ by adopting advanced memory management policies~\cite{aljundi2019gradient,buzzega2020rethinking}, exploiting meta-learning algorithms~\cite{riemer2018learning}, combining replay with knowledge distillation~\cite{rebuffi2017icarl,  buzzega2020dark}, or using the memory to train the model in an offline fashion~\cite{prabhu2020gdumb}.
\subsection{Semi-Supervised Learning}
\label{subsec:rel_ssl}
Semi-Supervised Learning studies how to improve supervised learning methods by leveraging additional unlabeled data. We exploit the latter in light of specific assumptions on how input and labels interact~\cite{olivier2006semi}. By assuming that close input data-points should correspond to similar outputs, \textbf{consistency regularization} encourages the model to produce consistent predictions for the same data-point. 
This principle can be applied either by comparing the predictions on the same exemplar by different learners~\cite{laine2016temporal,tarvainen2017mean} or the predictions on different augmentations of the same data-point by the same learner~\cite{berthelot2019mixmatch}.

Recently, several works investigated the refinement of such regularization through \textbf{adversarial training}, producing either more challenging perturbations~\cite{miyato2018virtual} or additional unsupervised samples for regularization purposes~\cite{zheng2017unlabeled}.

Our proposal, which we introduce in Sec.~\ref{sec:ccicc}, combines within-task consistency regularization with the dual strategy of maximizing cross-task feature dissimilarity. The latter reinforces deep representation learning according to the high-level structure of the target problem -- specifically, cross-task class disjunction. This can be seen as a form of Multi-Knowledge Representation~\cite{yang2021multiple} through the application of descriptive knowledge; on the other hand, our proposal remains open for further enrichment if additional knowledge on the target task were available~\cite{pan2020multiple}.

\section{Continual Semi-Supervised Learning}
\label{sec:semisup}

A supervised \textbf{Continual Learning} classification problem can be defined as a sequence $S$ composed of $T$ tasks. During each of the latter ($S = t\in\{1,...,T\}$), input samples $x$ and their corresponding ground truth labels $y$ are drawn from an i.i.d.\ distribution $D_t$. Considering a function $f$ with parameters $\theta$, we indicate its responses (logits) with $h_\theta(x)$ and the corresponding probability distribution over the classes with $f_\theta(x) \triangleq \argsoftmax(h_\theta(x))$. The goal is to find the optimal value for the parameters $\theta$ such that $f$ performs best on average on all tasks without incurring catastrophic forgetting; formally, we need to minimize the empirical risk over all tasks:
\begin{equation}
    \argmin_{\theta}\sum_{t=1}^{t_c}{\mathcal{L}_t},\text{where}~\mathcal{L}_t \triangleq {\mathds{E}_{(x,y) \sim D_t}\big[\ell(y, f_{\theta}(x))\big]}.
    \label{eq:obj}
\end{equation}
In \textbf{Continual Semi-Supervised Learning}, we propose to distribute the samples coming from $\mathcal{D}_t$ into two sets: $\mathcal{D}^s_t$, which contains a limited amount of pairs of labeled samples and their ground-truth labels ($x_s$, $y_s$) and $\mathcal{D}^u_t$, containing the rest of the unsupervised samples. We define this split according to a given proportion $p_s = \nicefrac{|\mathcal{D}^s_t|}{(|\mathcal{D}^s_t| + |\mathcal{D}^u_t|)}$ that remains fixed across all tasks. The objective of CSSL is optimizing Eq.~\ref{eq:obj} without having access to the ground-truth supervision signal for $\mathcal{D}^u_t$. Data from the stream consists of labeled pairs $\mathcal{S}\subset\mathcal{D}^s_t$ and unlabeled items $\mathcal{U}\subset\mathcal{D}^u_t$.

We are interested in shedding further light on CL models by understanding \textit{i)} how they perform under partial lack of supervision and \textit{ii)} how Semi-Supervised Learning approaches can be combined with them to exploit unsupervised data. Question \textit{i)} is investigated experimentally in Sec.~\ref{sec:exps_cl} and~\ref{sec:exps_clls} by evaluating methods that simply \textit{drop} unlabeled examples $x_u$. Differently, question \textit{ii)} opens up many possible solutions that we address by proposing \textbf{Contrastive Continual Interpolation Consistency} (CCIC).

\section{Method}
\label{sec:model}
We build our proposal upon two state-of-the-art approaches: on the one hand, we take advantage of \textbf{Experience Replay} (ER)~\cite{ratcliff1990connectionist,riemer2018learning} to mitigate catastrophic forgetting; on the other, we exploit \textbf{MixMatch}~\cite{berthelot2019mixmatch} to learn useful representations also from unlabeled examples. In the following: \textit{i)} to help the reader, we briefly recap the main traits of these algorithms (and let the original papers provide a deeper comprehension); \textit{ii)} we discuss how these two former approaches can be favorably complemented.
\subsection{Technical background}
\label{sec:bg}
As a first step, we equip the learner with a small memory buffer $\mathcal{M}$ (based on \textit{reservoir} sampling) and interleave a batch of examples drawn from it with each batch of the current task. Among all possible approaches, we opt for ER due to its lightweight design and effectiveness~\cite{riemer2018learning,buzzega2020rethinking}. 

When dealing with lack of supervision, \textit{self-training} represents a trivial strategy: here, the model itself produces the targets (\textit{pseudo-labels}) for unlabeled examples~\cite{yarowsky1995unsupervised, lee2013pseudo}. Unfortunately, this tends to become unstable with only a few annotations at disposal: as shown in our experiments, this encourages the model to overfit the limited supervised data available~\cite{oliver2018realistic}. 

Such an issue raises the need for a different objective, the latter being independent from the accuracy of the model on unlabeled examples. Consequently, we supplement our proposal with \textbf{MixMatch}~\cite{berthelot2019mixmatch}: the predictions of the network are not meant as training targets, but rather as means for applying \textit{consistency regularization}~\cite{tarvainen2017mean, miyato2018virtual}. Briefly, a soft-label is assigned to each unsupervised element by averaging and then sharpening the pre-softmax predictions of several different augmentations. 

To promote consistent responses to considerable variations of the data-points, labeled and unlabeled samples are combined through the mixUp procedure~\cite{zhang2017mixup}\footnote{Differently from MixMatch, we apply mixUp only on the input images (and not to the corresponding labels).}. Starting from the original sets $\mathcal{S}$ and $\mathcal{U}$ (respectively, labeled and unlabeled examples from the current batch), we thus obtain two final augmented and mixed sets of examples $\mathcal{S}^*$ and $\mathcal{U}^*$: in order to compute the loss terms $\mathcal{L}_{\mathtt{S}^*}$ and $\mathcal{L}_{\mathtt{U}^*}$, we use the ground truth labels for the examples of the former set and the soft-labels generated through response-averaging for the ones of the latter.
\begin{figure}[t]
    \centering
    \includegraphics[width=.99\linewidth]{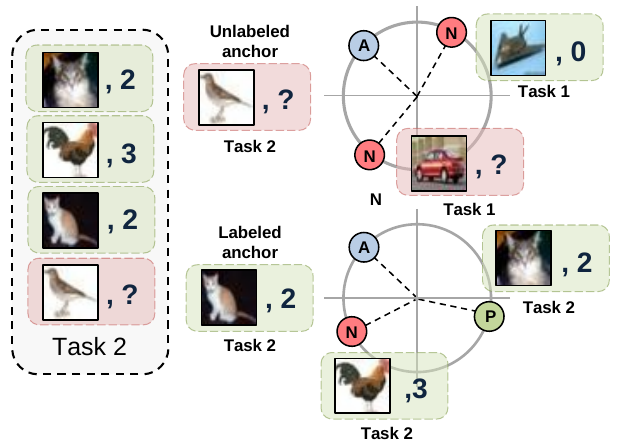} 
    \caption{CCIC exploits task identifiers to enforce semantic constraints: for each anchor (A), it asks the network to push away representations of different (N) tasks and move closer representations of the same one (P).}
    \vspace{-0.5em}
    \label{fig:mining}
\end{figure}
\subsection{Contrastive Continual Interpolation Consistency}
\label{sec:ccicc}
Supposing that boundaries between tasks are provided, we can associate in-memory exemplars with the task they come from. In the following, we discuss how this allows an additional weak form of supervision for unsupervised examples even if we do not know their classes exactly.

\vspace{0.5em}
\noindent\textbf{Unsupervised mining}. As tasks are disjoint, examples from different tasks necessarily belong to different classes: we account for that by adding a contrastive loss term, which pushes their responses away from each other (Fig.~\ref{fig:mining}). 
In details, we wish to maximize the Euclidean distance $D_{\theta}(x, x') \triangleq \left\lVert h_{\theta}(x) - h_{\theta}(x')\right\rVert^2_2$ between embeddings of examples of different tasks. Hence, we minimize:
\begin{equation}
    \mathcal{L}_\mathtt{UM} = \mathds{E}_{\substack{x \sim \mathcal{D}^{u}_{t_c}\\x_{N} \sim \mathcal{M}_{t < t_c}}}\big[ \operatorname{max}(\alpha - D_{\theta}(x, x_{N}), 0)\big],
    \label{eq:obj_10}
\end{equation}
where $t_c$ is the index of the current task $\mathcal{D}_{t_c}$, $\mathcal{M}_{t < t_c}$ indicates past examples from the memory buffer, and $\alpha$ is a constant margin beyond which no more efforts should be put into enlarging the distance between negative pairs. 

\vspace{0.5em}
\noindent\textbf{Supervised mining}. For each incoming labeled example, we also encourage the network to move its representation close to those belonging to the same class. We look for positive candidates $x_{P}$ within both the current batch and the memory buffer. In formal terms:
\begin{align}
\begin{aligned}
    \mathcal{L}_\mathtt{SM} = \mathds{E}_{x \sim \mathcal{D}^{s}_{t_c} \cup \mathcal{M}}\big[ \operatorname{relu}(\beta - D_{\theta}(x, x_{N}) + D_{\theta}(x, x_{P})\big].
    \label{eq:obj_11}
\end{aligned}
\end{align}

\vspace{0.5em}
\noindent\textbf{Overall objective}. To sum up, the objective of CCIC combines the consistency regularization term delivered by MixMatch with the two additional ones (Eq.~\ref{eq:obj_10} and Eq.~\ref{eq:obj_11}) applied in feature space; the overall optimization problem can be formalized as follows:
\begin{equation}\label{eq:loss_total}
    \operatorname{argmin}_{\theta}~~\mathcal{L} = \mathcal{L}_\mathtt{S} + \lambda\mathcal{L}_\mathtt{U} + \mathcal{L}_\mathtt{SM} + \mu\mathcal{L}_\mathtt{UM},
\end{equation}
where $\lambda$ and $\mu$ are hyperparameters setting the importance of the unsupervised examples.

\vspace{0.5em}
\noindent\textbf{Exploiting distance metric learning during inference}. Once we have introduced constraints in feature space (Eq.~\ref{eq:obj_10},~\ref{eq:obj_11}), we can also exploit them by devising a different inference schema, which further contributes to relieve forgetting. Similarly to~\cite{rebuffi2017icarl}, we employ the k-Nearest Neighbors algorithm as final classifier, thus decoupling classification from feature extraction. This has been shown beneficial in Continual Learning, as it saves the final fully-connected layer from continuously keeping up with the changing features (and \textit{vice versa}). As kNN is non-parametric and builds upon the feature space solely, it fits in harmony with the rest of the model, controlling the damage caused by catastrophic forgetting. We fit the kNN classifier using the examples of memory buffer as training set.
\section{Experiments}
\label{sec:exps}
\begin{table*}[t]
    \centering
    \caption{Average Accuracy of CL Methods and of Our Proposals on CSSL Benchmarks.}
    \sisetup{detect-weight, mode=math, table-format = 2.2}
\resizebox{\textwidth}{!}{
\setlength{\tabcolsep}{0.45em}
\begin{tabular}{l!{\textbrokenbar}ccc!{·}c!{\textbrokenbar}ccc!{·}c!{\textbrokenbar}ccc!{·}c}
\toprule
\textbf{Class-IL}   & \multicolumn{4}{c!{\textbrokenbar}}{\textbf{SVHN} (UB: \shnormres{86.18}{1.75})}                                                            & \multicolumn{4}{c!{\textbrokenbar}}{\textbf{CIFAR-10} (UB: \shnormres{92.12}{0.079})} &\multicolumn{4}{c}{\textbf{CIFAR-100} (UB: \shnormres{67.71}{0.91})}                                       \\
\midrule
\small{Labels \%} & $\mathit{0.8}\%$ & $\mathit{5}\%$ & $\mathit{25}\%$ & $\mathit{100}\%$ & $\mathit{0.8}\%$ & $\mathit{5}\%$ &  $\mathit{25}\%$ & $\mathit{100}\%$& $\mathit{0.8}\%$ & $\mathit{5}\%$  & $\mathit{25}\%$ & $\mathit{100}\%$\\
\midrule
         Fine Tuning         &   \shnormres{9.9}{1.67}    &   \shnormres{9.93}{8.38}  &     \shnormres{17.53}{9.38}    &  \shnormres{17.76}{1.21}  &    \shnormres{13.55}{2.91}   &   \shnormres{18.20}{0.41}   &      \shnormres{19.18}{2.23}    &    \shnormres{19.63}{8.43}   &     \shnormres{1.81}{0.15}    &     \shnormres{4.95}{0.28}    &     \shnormres{7.82}{0.10}      & \shnormres{8.60}{0.36} \\    
         LwF         &     \shnormres{9.93}{0.3}     &   \shnormres{9.93}{1.85}  &     \shnormres{14.82}{3.62}    &  \shnormres{16.88}{0.09}  &    \shnormres{13.07}{2.21}   &   \shnormres{17.67}{3.17}   &      \shnormres{19.38}{1.7}    &   \shnormres{19.61}{10.3}  &     \shnormres{1.58}{0.07}    &     \shnormres{4.49}{0.14}    &      \shnormres{8.00}{0.12}     & \shnormres{8.43}{0.45} \\    
         oEWC        &     \shnormres{9.93}{0.23}     &   \shnormres{9.93}{0.74}  &     \shnormres{14.70}{0.48}    &  \shnormres{17.94}{0.19}  &   \shnormres{13.70}{1.16}  &  \shnormres{17.55}{1.23}  &   \shnormres{19.1}{0.81}  &   \shnormres{19.55}{7.5}  &   \shnormres{1.42}{0.1}  &   \shnormres{4.67}{0.11}  &   \shnormres{7.83}{0.4}    & \shnormres{7.79}{0.13}  \\     
          SI         &     \shnormres{9.93}{1.19}     &  \shnormres{10.22}{5.86}  &     \shnormres{17.13}{7.65}    &  \shnormres{18.18}{0.17}  &   \shnormres{12.40}{0.39}  &  \shnormres{15.94}{0.98}  &     \shnormres{19.24}{1.34}   &   \shnormres{19.51}{3.32}  &     \shnormres{1.29}{0.2}    &     \shnormres{3.40}{0.19}    &      \shnormres{7.47}{0.52}     & \shnormres{8.11}{1.16}  \\     
\midrule
      ER$_{500}$     &    \shnormres{32.50}{7.08}     &  \shnormres{56.04}{1.97}  &     \shnormres{59.65}{1.78}    &  \shnormres{66.50}{2.78}  &    \shnormres{36.27}{1.12}   &   \shnormres{51.88}{4.5}   &      \shnormres{60.92}{5.7}    &    \shnormres{62.19}{2.6}   &   \shnormres{8.23}{0.14}  &  \shnormres{13.67}{0.58}  &   \shnormres{17.05}{0.71}   & \shnormres{21.30}{0.24}  \\
    iCaRL$_{500}$    &     \shnormres{8.90}{0.36}     &  \shnormres{10.04}{1.54}  &     \shnormres{19.92}{1.23}    &  \shnormres{23.09}{2.44}  &   \shnormres{24.74}{2.33}  &  \shnormres{35.79}{3.23}  &     \shnormres{51.36}{8.4}   &  \shnormres{60.96}{0.43}  &   \shnormres{3.61}{0.06}  &   \shnormres{11.26}{0.25}  &  \shboldres{27.60}{0.36}     & \shboldres{37.81}{0.27}  \\  
     DER$_{500}$     &    \shnormres{11.88}{1.66}     &  \shnormres{54.56}{2.63}  &     \shnormres{56.86}{5.84}   &  \shboldres{70.81}{3.73}  &    \shnormres{29.06}{0.36}   &   \shnormres{35.31}{8.3}   &      \shnormres{49.95}{2.27}    &    \shboldres{67.10}{1.63}   &   \shnormres{1.66}{0.05}  &   \shnormres{5.10}{0.85}  &   \shnormres{13.03}{5.28}   & \shnormres{28.77}{7.23}\\  
    GDumb$_{500}$    &      \shnormres{34.58}{5.13}       &    \shnormres{41.76}{8.3}    &      \shnormres{59.20}{8.5}       &  \shnormres{59.91}{9.7}      &    \shnormres{39.55}{9.55}    &   \shnormres{40.85}{11.76}    &      \shnormres{44.80}{5.35}     & \shnormres{47.90}{1.62}       &     \shnormres{8.58}{0.12}    &     \shnormres{9.94}{0.38}    &   \shnormres{10.13}{0.35}   & \shnormres{10.95}{1.84}\\  
    PseudoER$_{500}$ &   \shnormres{23.2}{0.74}   &   \shnormres{48.93}{1.17} &    \shnormres{63.55}{2.74}     &             -    &  \shnormres{37.79}{1.63} &  \shnormres{44.87}{2.33} &  \shnormres{56.28}{1.63} & -                                                                                                                   &   \shnormres{5.14}{0.57}  &  \shnormres{14.33}{0.06}  & \shnormres{18.45}{0.45}       &  - \\
     CCIC$_{500}$    &   \shboldres{55.3}{3.21}   &   \shboldres{70.06}{3.87}  &      \shboldres{75.86}{1.51}    &             -             &   \shboldres{53.96}{0.2}  &  \shboldres{63.29}{1.9}   &     \shboldres{63.86}{2.6}   &             -             &  \shboldres{11.53}{0.73}  &  \shboldres{19.52}{0.19}  &     \shnormres{20.33}{0.30}  &  - \\  
\midrule
     ER$_{5120}$     &    \shnormres{44.37}{1.44}     &  \shnormres{69.94}{3.55}  &     \shnormres{77.57}{8.66}    &  \shboldres{80.54}{3.22}  &    \shnormres{37.37}{2.3}   &   \shnormres{64.10}{5.3}   &      \shnormres{79.66}{1.2}    &    \shnormres{83.28}{2.8}   &   \shnormres{9.61}{0.60}  &  \shnormres{22.83}{0.30}  &   \shnormres{37.85}{0.57}   & \shboldres{48.95}{0.18}  \\  
    iCaRL$_{5120}$   &     \shnormres{9.29}{0.24}     &  \shnormres{11.54}{0.52}  &     \shnormres{19.53}{3.73}    &  \shnormres{23.89}{4.45}  &   \shnormres{20.73}{3.32}  &  \shnormres{35.47}{5.6}  &     \shnormres{56.31}{2.19}   &  \shnormres{61.90}{1.53}  &   \shnormres{4.26}{0.06}  &  \shnormres{12.18}{0.30}  &   \shnormres{30.88}{1.01}   & \shnormres{41.15}{0.43}  \\     
     DER$_{5120}$    &    \shnormres{23.10}{0.96}     &  \shnormres{67.75}{5.17}  &     \shnormres{74.72}{2.35}    &  \shnormres{75.32}{7.57}  &    \shnormres{32.91}{0.9}   &   \shnormres{47.58}{2.21}   &      \shnormres{73.93}{4.5}    &    \shboldres{84.49}{2.1}   &   \shnormres{1.58}{0.07}  &   \shnormres{4.65}{0.55}  &   \shnormres{11.93}{3.38}   & \shnormres{38.59}{3.64} \\  
    GDumb$_{5120}$   &      \shnormres{46.54}{8.04}       &    \shnormres{74.38}{2.25}    &     \shnormres{74.63}{3.78}        &   \shnormres{78.27}{2.25} &    \shnormres{40.76}{12.65}    &   \shnormres{71.19}{2.62}    &      \shnormres{81.37}{0.78}     &   \shnormres{82.53}{0.54}     &   \shnormres{9.63}{1.11}  &  \shnormres{23.26}{0.08}  &   \shnormres{33.24}{2.23}   & \shnormres{42.92}{1.73}\\  
    PseudoER$_{5120}$&   \shnormres{45.8}{2.79}   &   \shnormres{74.62}{2.43} &    \shnormres{77.85}{0.82}     &             -&  \shboldres{62.22}{2.06}&  \shnormres{72.88}{2.02} &  \shnormres{80.37}{0.13} & -                                                                                                                   &   \shnormres{8.18}{1.39}  &  \shnormres{25.12}{1.57}  & \shnormres{40.04}{0.42}       &  - \\
    CCIC$_{5120}$    &  \shboldres{59.31}{5.30}  &   \shboldres{80.99}{2.26}  &     \shboldres{83.93}{0.21}    &             -             &   \shnormres{55.19}{1.4}  &  \shboldres{74.34}{1.7}  &     \shboldres{84.74}{0.9}   &             -             &  \shboldres{12.02}{0.30}  &  \shboldres{29.54}{0.37}  &    \shboldres{44.28}{0.07}  & - \\

\bottomrule
\end{tabular}
}
\vspace{-0.7em}
    \label{tab:resssti}
\end{table*}
We conduct our experiments on three standard datasets\footnote{Code available at \url{https://github.com/loribonna/CSSL}.}. \textbf{Split SVHN}: five subsequent binary tasks built on top of the Street View House Numbers (SVHN) dataset~\cite{netzer2011reading}; \textbf{Split CIFAR-10}: equivalent to the previous one, but using the CIFAR-10 dataset~\cite{krizhevsky2009learning}. \textbf{Split CIFAR-100}: a longer and more challenging evaluation in which the model is presented ten subsequent tasks, each comprising of $10$ classes from the CIFAR-100 dataset~\cite{krizhevsky2009learning}.

We vary the fraction of labeled data shown to the model ($p_s$) to encompass different degrees of supervision (${0.8}\%$, ${5}\%$, ${25}\%$, and ${100}\%$, \textit{i.e.}, $400$, $2500$, $25000$, and $50000$ samples for CIFAR-10/100). For fairness, we keep the original balancing between classes in both train and test sets; in presence of low rates, we make sure that each class is represented by a proportional amount of labels.
\minispace
\noindent\textbf{Architectures.}~As in~\cite{abati2020conditional}, experiments on Split SVHN are conducted on a small CNN, comprising of three ReLU layers interleaved by max-pooling. Instead, we rely on ResNet18 for CIFAR-10 and CIFAR-100, as done in~\cite{buzzega2020dark}.

\minispace
\noindent\textbf{Metrics.}~We report the performance in terms of average final accuracy, as done in~\cite{lopez2017gradient, aljundi2019gradient}. 
Accuracies are averaged across 5 runs (we also report standard deviations). 
\minispace
\noindent\textbf{Implementation details.}~As discussed in Sec.~\ref{sec:model}, our proposals rely on data augmentation to promote consistency regularization. We apply random cropping and horizontal flipping (except for Split SVHN); the same choice is applied to competitors to ensure fairness. To perform hyperparameters selection (learning rate, batch size, optimization algorithm, and regularization coefficients), we perform a grid search on top of a validation set (corresponding to $10\%$ of the training set), as done in~\cite{riemer2018learning,buzzega2020dark,rebuffi2017icarl}. For CCIC, we keep the number of augmentations fixed to $3$ and report chosen values for $\lambda$ and $\mu$ in Tab.~\ref{tab:hp}.
To guarantee fairness, we fix the batch size and memory minibatch size to $32$ for all models. We train on each task for $10$ epochs on SVHN, for $50$ on CIFAR-10, and $30$ on CIFAR-100. All methods use SGD as an optimizer with the only exception of CCIC, which employs Adam.
\begin{table}[t]
    \centering
    \vspace{-0.5em}
    \caption{Values of $(\lambda,\mu)$ for CCIC chosen after the grid-search.}
    \begin{tabular}{llccc}
    \toprule
         \textbf{Lab.\ \%} & \textbf{$|\mathcal{M}|$} & \textbf{SVHN} & \textbf{CIFAR-10} & \textbf{CIFAR-100} \\
        \bottomrule
        \multirow{2}{*}{$\mathit{0.8}\%$} & $500$  & $(0.5, 0.5)$ & $(0.5, 0.5)$ & $(0.3, 1.0)$\\
                         & $5120$ & $(0.1, 0.5)$ & $(0.5, 0.5)$ & $(0.3, 0.3)$\\
        \hdashline
        \multirow{2}{*}{$\mathit{5}\%$}   & $500$  & $(0.1, 0.5)$ & $(0.3, 0.5)$ & $(0.5, 0.5)$\\
                         & $5120$ & $(0.1, 0.5)$ & $(0.5, 0.5)$ & $(0.5, 0.5)$\\
        \hdashline
        \multirow{2}{*}{$\mathit{25}\%$}  & $500$  & $(0.5, 0.5)$ & $(0.1, 0.5)$ & $(0.5, 0.7)$\\
                         & $5120$ & $(0.5, 0.5)$ & $(0.1, 1.0)$ & $(0.5, 0.5)$\\
    \bottomrule
    \end{tabular}
    \vspace{-0.5em}
    \label{tab:hp}
\end{table}
\subsection{Baselines}
\label{sec:exps_cl}
\noindent\textbf{Lower/Upper bounds}.~We bound the performance for our experiments by including two reference measures. As a lower bound, we evaluate the performance of a model trained by \textit{Fine Tuning} exclusively on the set of supervised examples, without any countermeasure to catastrophic forgetting. We also provide an upper-bound (UB) given by a model trained jointly, \textit{i.e.}, without dividing the dataset into tasks or discarding any ground-truth annotation.
\minispace
\noindent\textbf{Drop-the-unlabeled}.~The most straightforward approach to adapt existing methods to our setting consists in simply discarding unlabeled examples from the current batch. In this regard, we compare our proposal with Learning Without Forgetting (LwF)~\cite{li2017learning}, online Elastic Weight Consolidation (oEWC)~\cite{schwarz2018progress}, Synaptic Intelligence (SI)~\cite{zenke2017continual}, Experience Replay (ER)~\cite{riemer2018learning}, iCaRL~\cite{rebuffi2017icarl}, Dark Experience Replay (DER)~\cite{buzzega2020dark} and GDumb~\cite{prabhu2020gdumb}. By so doing, we can verify whether our proposal is able to better sustain a training regime with reduced supervision.
\minispace
\noindent\textbf{Pseudo-Labeling}.~Inspired by the line of works relying on self-labeling~\cite{yarowsky1995unsupervised,lee2013pseudo}, we here introduce a simple CSSL baseline that allows ER to profit from the unlabeled examples: given
an unlabeled example $x_u$, it pins as a \textit{pseudo-label} $\tilde{y}_u$~\cite{lee2013pseudo} the prediction of the model itself. Formally, 
\begin{equation}
    \tilde{y}_u = \operatorname{argmax}_{c\in C_t} \hspace{0.3em} h^c_\theta(x_u),
\end{equation}
where $C_t$ is the set of classes of the current task. As discussed in Sec.~\ref{sec:bg}, self training is likely to cause model instability (especially at task boundaries, when the model starts to experience new data): we mitigate this by applying a threshold $\eta$ to discard low-confidence outputs and their relative $x_u$. Specifically, we estimate the confidence as the difference between the two highest values of $h^c_\theta(x_u)$. After this step, a pair $(x_u, \tilde{y}_u)$ is considered on a par with any supervised pair $(x_s, y_s)$, and is therefore inserted into the memory buffer. We refer to this baseline as \textbf{PseudoER}.
\subsection{Experimental Results}
\label{sec:exps_clls}
As revealed by the results in Tab.~\ref{tab:resssti}, CSSL proves to be a challenging scenario. Unsurprisingly, its difficulty increases when fewer labels are provided to the learner.
\minispace
\noindent\textbf{Regularization methods} are generally regarded as weak in the Class-IL scenario~\cite{farquhar2018towards,aljundi2019gradient}. This conforms with our empirical observations, as LwF, oEWC and SI underperform across all datasets. Indeed, these methods rarely outperform our lower bound (Fine Tuning), indicating that they are not effective outside of Task-IL and Domain-IL. This becomes especially evident in the low-label regime.
\minispace
\noindent\textbf{Rehearsal methods} overall show an expected decrease in performance as supervision diminishes. This is especially severe for DER and iCaRL, as their accuracy drops on average by more than $70\%$ between $100\%$ and $0.8\%$ labels. As the model underfits the task when less supervision is provided, it produces less reliable targets that cannot be successfully used for replay by these methods. In contrast, ER is able to replay information successfully as it exploits hard targets; thus, it learns effectively even after initially underfitting the task. Indeed, its accuracy with $5\%$ labels and buffer $5120$ is always higher than its fully-supervised accuracy with a smaller buffer. While ER is able to overcome the lack of labels when paired with an appropriate buffer, knowledge-distillation based approaches remarkably encounter a major hindrance in this setting.

We attribute the failure of iCaRL on SVHN to the low complexity of the backbone network. Indeed, a shallow backbone provides for a latent space that is less suitable for its nearest-mean-of-exemplars classifier. Conversely, this method proves quite effective even with a reduced memory buffer on CIFAR-100. In this benchmark, the \textit{herding sampling} of iCaRL ensures that all classes are fairly represented even in a small memory buffer.

Finally, GDumb does not suffer from lower supervision as long as its buffer can be filled completely: its operation is not disrupted by unlabeled examples on the stream, as it ignores the latter entirely. While it outperforms other methods when few labels are available, CCIC surpasses it consistently. This suggests that the stream offers potential for further learning and should not be dismissed.
\minispace
\noindent\textbf{CSSL Methods}.~Our PseudoER baseline performs notably well on CIFAR-10, maintaining high accuracy as the amount of supervision decreases. However, while CIFAR-10 is a complex benchmark, it only features two classes for each task, which makes it easy for pseudo-labeling to produce reasonable responses (it is noted that a random guess would result in $50\%$ accuracy).
Conversely, PseudoER struggles to produce valid targets and exhibits a swift performance drop on CIFAR-100 as the availability of labeled data decreases. Similarly, we find the application of pseudo-labeling beneficial for SVHN only as the space reserved for the buffer increases, demonstrating the pitfalls of this approach in the online setting.

On the contrary, the compelling performance of CCIC indicates successful blending of supervised information and semi-supervised regularization. While ER encounters an average performance drop of $47\%$, going from $25\%$ to $0.8\%$ labels on CIFAR-10, CCIC only loses $26\%$ on average.
Surprisingly, we observe that -- for the majority of evaluated benchmarks -- $25\%$ supervision is enough to approach the results of fully-supervised methods, even outperforming the state-of-the-art in some circumstances (CIFAR-10 with buffer size $5120$, SVHN with buffer size $500$ and $5120$).

This hints that, when learning from a stream of data, striving to provide full supervision is not as essential as it might be expected: differently from the offline scenario, a greater amount of labels might not produce a proportionate profit due to \textit{catastrophic forgetting}.
In this respect, our experiments suggest that pairing few labeled examples with semi-supervised techniques represents a more efficient paradigm to achieve satisfying performance.
\begin{table}[t]
\centering
\caption{Unsupervised Mining Techniques for CCIC on CIFAR-100.}
\begin{tabular}{lcc}
\toprule
\textbf{Labels \%~($|\mathcal{M}|=5120$)}  & $\mathit{5\%}$ & $\mathit{25\%}$  \\
\midrule
     Across-Task Mining (Eq.~\ref{eq:obj_10})  & \shnormres{29.5}{0.4}  & \shnormres{44.3}{0.1} \\
     Within-Task Mining   & \shnormres{29.25}{0.15}& \shnormres{44.02}{0.23}  \\   
     Task-Agnostic Mining & \shnormres{29.06}{0.70}& \shnormres{43.91}{0.76}  \\    
\bottomrule
\end{tabular}
\vspace{-0.8em}
\label{tab:minings}
\end{table}
\minispace
\noindent\textbf{Unsupervised Mining in CCIC.} In its unsupervised mining loss term $\mathcal{L}_\mathtt{UM}$, CCIC takes examples of previous tasks in the memory buffer as negatives (\textit{Across-Task Mining}) and requires their representations to be pushed away from current data.
In Tab.~\ref{tab:minings}, we compare this design choice with two alternative strategies: \textit{i)} \textit{Within-Task Mining}, where we let the model choose the negatives from the current task only; and \textit{ii)}
\textit{Task-Agnostic Mining}, where the model can freely pick a negative example from either the memory or the current batch without any task-specific prior. As can be observed, Task-Agnostic Mining and Within-Task Mining lead to a small but consistent decrease in performance, while $\mathcal{L}_\mathtt{UM}$ proves to be the most rewarding strategy.
\begin{table}[t]
    \centering
    \caption{Average Accuracy of alternative CSSL proposals on CIFAR-10}
    \label{tab:ema_abla}
    \begin{tabular}{lccc}
\toprule
\textbf{Labels \%}  & $\mathit{0.8\%}$ & $\mathit{5\%}$ & $\mathit{25\%}$  \\
\midrule
ER+EMA$_{500}$ &   \shnormres{21.37}{0.54} &  \shnormres{26.31}{1.02} &  \shnormres{43.25}{1.22} \\
CCIC$_{500}$ &   \shnormres{53.96}{0.2} &   \shnormres{63.29}{2.1} &   \shnormres{63.86}{2.6} \\
\midrule
ER+EMA$_{5120}$ &  \shnormres{25.85}{0.80} &  \shnormres{40.77}{2.08} &  \shnormres{64.75}{0.42} \\
CCIC$_{5120}$ &    \shnormres{55.19}{1.2} &   \shnormres{74.34}{1.7} &   \shnormres{84.74}{0.9} \\
\bottomrule
\end{tabular}
\vspace{-0.6em}
\end{table}
\minispace
\noindent\textbf{Model-driven Consistency.} In addition with combining a contrastive form of consistency regularization with ER, we propose an additional \textit{temporal consistency} baseline which requires the activations of the model to match a slower moving-average checkpoint. Results in Tab.~\ref{tab:ema_abla} show, however, that such approach under-performs consistently, not even reaching the performance of ER. This suggests that, differently from fully-supervised scenarios~\cite{tarvainen2017mean}, exponential moving average approaches do not necessarily scale to CL.
\section{Conclusion}
\label{sec:conclusions}
Catastrophic forgetting prevents most current state-of-the-art models from sequentially learning multiple tasks, forcing practitioners to heavy resource-demanding training processes.
Moreover, many of the applications that might benefit from CL algorithms are often characterized by label scarcity. For this reason, we investigate the possibility of leveraging unlabeled data-points to enhance the performance of Continual Learning models, a scenario that we name \textbf{Continual Semi-Supervised Learning (CSSL)}.

We further propose \textbf{Constrastive Continual Interpolation Consistency (CCIC)}, an incremental approach that combines the benefits of rehearsal with consistency regularization and distance-based constraints. Remarkably, our experiments suggest that well-designed methods can effectively exploit the unlabeled examples to prevent forgetting. This indicates that the effort of annotating all data may be unnecessary in a continual scenario.
%

\vspace{-0.9em}
\section*{Acknowledgement}
This work was supported by the FF4EuroHPC: HPC Innovation for European SMEs, Project Call 1. Project FF4EuroHPC has received funding from the European High-Performance Computing Joint Undertaking (JU) under grant agreement No 951745.
\vspace{-0.9em}
\bibliography{references.bib}

\end{document}